\def\CLASSINPUTinnersidemargin{0.75in}
\def\CLASSINPUToutersidemargin{0.75in}
\def\CLASSINPUTtoptextmargin{0.82in}
\def\CLASSINPUTbottomtextmargin{0.75in}
\documentclass[conference]{IEEEtran} 

\IEEEoverridecommandlockouts                              

\usepackage[T1]{fontenc}

\usepackage{amsmath}
\usepackage{amssymb}
\usepackage{mathtools}
\usepackage{mathrsfs}

\usepackage{graphicx}
\graphicspath{{./images/}}

\usepackage{booktabs}
\usepackage{multirow}
\usepackage{tabularx}
\usepackage{array}
\usepackage{adjustbox}
\usepackage{colortbl}
\usepackage{subcaption}

\usepackage{enumitem}
\usepackage[normalem]{ulem}
\usepackage{pifont}
\newcommand{\xmark}{\ding{55}}
\usepackage{gensymb}
\usepackage{siunitx}
\usepackage{interval}

\usepackage{stfloats}
\usepackage{pdflscape}
\usepackage{afterpage}

\usepackage[ruled,vlined]{algorithm2e}

\usepackage{tikz}

\usepackage{cite}

\usepackage{url}
\usepackage[breaklinks,colorlinks,pagebackref]{hyperref}

\newcommand\copyrighttext{%
  \footnotesize This work has been accepted for publication at the 2026 IEEE 29th International Conference on Intelligent Transportation Systems (ITSC).\\
  \textcopyright 2026 IEEE. Personal use of this material is permitted. Permission from IEEE must be obtained for all other uses, in any current or future media, including reprinting/republishing this material for advertising or promotional purposes, creating new collective works, for resale or redistribution to servers or lists, or reuse of any copyrighted component of this work in other works.
}
  
\newcommand\copyrightnotice{%
\begin{tikzpicture}[remember picture,overlay]
\node[anchor=south,yshift=5pt] at (current page.south) {\fbox{\parbox{\dimexpr\textwidth-\fboxsep-\fboxrule\relax}{\copyrighttext}}};
\end{tikzpicture}
}

\title{\vspace*{18pt}\LARGE \bf
Invascal: Inverse-Vacuity Self-Calibration for Uncertainty-Aware LiDAR Range-View Semantic Segmentation
}
\author{Kerim Turacan, Hannes Reichert, Andrei Bolandut and Konrad Doll
	\thanks{K. Turacan, H. Reichert, A. Bolandut and K. Doll are with the Faculty of Engineering and Computer Science,
		University of Applied Sciences Aschaffenburg, Aschaffenburg, Germany
		{\tt\footnotesize firstname.lastname@th-ab.de}}
}

\makeatletter
\def\bstctlcite{\@ifnextchar[{\@bstctlcite}{\@bstctlcite[@auxout]}}
\def\@bstctlcite[#1]#2{\@bsphack
  \@for\@citeb:=#2\do{%
    \edef\@citeb{\expandafter\@firstofone\@citeb}%
    \if@filesw\immediate\write\csname #1\endcsname{\string\citation{\@citeb}}\fi}%
  \@esphack}
\makeatother

\begin{document}
\bstctlcite{IEEEexample:BSTcontrol}

\maketitle

\copyrightnotice
\thispagestyle{empty}
\pagestyle{empty}


\begin{abstract}
LiDAR semantic segmentation is a core perception capability for autonomous vehicles and mobile robots. However, safe operation also depends on knowing when predictions are unreliable. 
Existing approaches typically rely on softmax confidence, which is often miscalibrated and overconfident, while stronger uncertainty estimates from Monte Carlo dropout or ensembles are often computationally expensive for real-time use.
To this end, we introduce a novel, architecture-agnostic uncertainty-aware Adapter Head. It decomposes the prediction into a Preference Head for class ranking and a Strength Head that refines uncertainty assessment, thereby enabling a principled construction of evidential Dirichlet representations. Building on this design, we propose our inverse-vacuity self-calibration objective (Invascal), which directly supervises the strength signal to produce reliable and well-calibrated uncertainty estimates while preventing runaway evidence growth.
We evaluate our framework across multiple LiDAR datasets and backbone architectures. We compare against deterministic training, Monte Carlo dropout and ensembles, and prior evidential methods.
Our approach consistently improves uncertainty calibration over traditional deterministic methods with minimal computational overhead. 
At the same time, it preserves competitive segmentation accuracy, where prior evidential methods often suffer performance degradation.
The code is available at \url{https://github.com/kav-institute/LidarRV-SemSeg-Unc}.
\end{abstract}



\section{\large Introduction}
\label{sec_introduction}

LiDAR provides accurate, illumination-invariant geometry over long ranges, making it a strong complement to cameras for autonomous vehicles and other safety-critical robotic systems. However, dense semantic understanding from LiDAR is challenging -- point clouds are sparse and irregular, density falls with range, and occlusions complicate local reasoning -- and safe operation also requires knowing when predictions cannot be trusted.
LiDAR semantic segmentation methods convert point clouds into structured representations -- point-based \cite{thomas2019kpconvflexibledeformableconvolution, wu2022pointtransformerv2grouped, wu2024pointtransformerv3simpler}, sparse voxel \cite{zhu2020cylindricalasymmetrical3dconvolution, hou2022pointtovoxelknowledgedistillationlidar}, bird's-eye view (BEV) \cite{zhang2020polarnetimprovedgridrepresentation, sirohi2021efficientlpsefficientlidarpanoptic}, or range-view \cite{yang2025flaresfastaccuratelidar, milioto2019rangenetpp, SalsaNext, reichert2025realtimesemanticsegmentation} -- trading geometric fidelity against compute and memory. 
We focus on range-view, whose 2D projection preserves measurement ordering and a near one-to-one correspondence between pixels and LiDAR rays, making it well suited to real-time deployment.
BEV methods provide a useful metric structure and are effective for downstream grouping tasks \cite{zhang2020polarnetimprovedgridrepresentation, sirohi2021efficientlpsefficientlidarpanoptic}, but discretization introduces artifacts through point pooling, vertical collapse, and label aggregation heuristics \cite{sirohi2022uncertaintyawarelidarpanopticsegmentation, zhou2021panopticpolarnetproposalfreelidarpoint}. 
In autonomous driving, where perception must cover hundreds of meters, metric voxelization scales poorly and forces aggressive point pooling to keep computation tractable. 
Range-view instead orders returns on the spherical projection's angular grid and can operate at the sensor's native resolution. The model must therefore commit per ray, so confidence attaches directly to a physical measurement with no aggregation step to absorb ambiguity.
The trade-off is that image-plane neighbors can lie on physically distant points, and the spherical projection distorts distant returns into limited pixel coverage.

In practice, uncertainty is often approximated by softmax confidence, which is known to be miscalibrated and frequently overconfident \cite{guo2017calibrationmodernneuralnetworks}.
Monte Carlo dropout and deep ensembles improve calibration but rely on repeated forward passes or replicated models that erode the latency advantage of range-view.
Evidential deep learning (EDL) \cite{sensoy2018evidentialdeeplearningquantify} yields closed-form uncertainty in a single pass, but existing evidential LiDAR work targets BEV panoptic segmentation and open-set extensions \cite{sirohi2022uncertaintyawarelidarpanopticsegmentation, mohan2025opensetlidarpanopticsegmentation}.
LiDAR range-view semantic segmentation lacks a comparable controlled study of Dirichlet-family losses. 
Moreover, classical EDL couples class ranking and confidence in the same per-class outputs and typically relies on Kullback-Leibler regularization to suppress runaway evidence, which we show can substantially degrade segmentation accuracy on safety-relevant classes.

We therefore propose an evidential range-view formulation that produces uncertainty in a single forward pass, controls confidence independently of class ranking, and supervises evidence strength via a self-calibration target tied to the model's own correctness rather than indirectly via off-class regularization. 
This decoupling of confidence from class ranking preserves the accuracy of the deterministic baseline while notably improving entropy-based uncertainty calibration.

Our contributions are:
\begin{enumerate}
\item A novel, architecture-agnostic uncertainty-aware Adapter Head using an EDL-inspired Dirichlet reparameterization to decouple class preference and evidence strength.
\item An inverse-vacuity self-calibration objective (Invascal) that supervises the Strength Head to produce reliable and well-calibrated uncertainty estimates.
\item A comprehensive evaluation of segmentation accuracy and calibration to characterize reliability trade-offs in LiDAR semantic segmentation.
\end{enumerate}

\section{\large Related Work}
\label{sec_sota}

Uncertainty estimation for semantic segmentation follows two main directions.
The first uses sampling-based approximations to Bayesian inference, such as Monte Carlo dropout or ensembles, to attach uncertainty estimates to standard decoders.
The second uses sampling-free distributional models, most commonly Dirichlet-based evidential approaches, which predict distribution parameters and yield closed-form uncertainty in a single forward pass.
Our work belongs to the latter category.
We review both 2D image and LiDAR segmentation literature, focusing on methods that provide dense uncertainty estimates.

\subsection{Sampling-based uncertainty for semantic segmentation}
Bayesian SegNet \cite{kendall2016bayesiansegnetmodeluncertainty} demonstrated pixel-wise uncertainty estimation using stochastic dropout inference in encoder-decoder networks.
Kendall and Gal \cite{kendall2017uncertaintiesneedbayesiandeep} combined MC-dropout for epistemic uncertainty with a learned heteroscedastic likelihood for aleatoric uncertainty.
Probabilistic U-Net \cite{kohl2019probabilisticunetsegmentationambiguous} modeled a distribution over plausible segmentations to capture structured ambiguity by using a conditional variational autoencoder on top of U-Net.

In LiDAR segmentation, uncertainty estimation initially followed a similar path.
SalsaNext introduced Bayesian uncertainty estimation for range-view LiDAR segmentation on the SemanticKITTI dataset\cite{SemanticKITTI_behley2019iccv}.
More recent Bayesian point-cloud segmentation studies confirm the trade-off between uncertainty quality and runtime efficiency \cite{LiDARsegm_BNN_Vassilev_2024}, motivating the development of sampling-free alternatives for real-time systems.

\subsection{Confidence learning for failure prediction and OOD detection}
A related sampling-free line of work trains auxiliary confidence estimators for failure prediction or out-of-distribution (OOD) detection \cite{devries2018learningconfidenceoutofdistributiondetection,corbiere2019addressingfailurepredictionlearning,corbiere2021confidenceestimationauxiliarymodels,rahman2021fsnetfailuredetectionframework,hendrycks2022scalingoutofdistributiondetectionrealworld}.
LiDAR semantic segmentation variants of this approach have also been proposed \cite{miandashti2025calibratedefficientsamplingfreeconfidence,shojaei2024uncertaintyestimationoutofdistributiondetection}.
These methods produce useful confidence scores for rejection or screening, but they do not explicitly model a calibrated predictive distribution over classes.
Evidential methods instead parameterize the predictive distribution itself, which enables uncertainty measures tied to the distribution parameters.

\subsection{Dirichlet and evidential segmentation}
Evidential Deep Learning (EDL) \cite{sensoy2018evidentialdeeplearningquantify} predicts Dirichlet parameters that can be interpreted as class evidence, enabling single-pass uncertainty estimation.
While early work focused on classification benchmarks such as MNIST \cite{MNIST_2012} and CIFAR-10 \cite{CIFAR-10_dataset}, evidential ideas were extended to semantic segmentation and OOD detection.
Recent examples combine Dirichlet outputs with intermediate-layer variational inference to mitigate overconfidence \cite{DirchletDNN-ILVI_ICCVW2023} and \cite{EvidentialSemanticSegmOOD_Ancha_ICRA2024,charpentier2020posteriornetworkuncertaintyestimation}
augment segmentation heads with feature-space density models to improve OOD obstacle detection.

In LiDAR perception, evidential modeling has been explored mainly in BEV/polar-BEV panoptic segmentation.
EvLPSNet \cite{sirohi2022uncertaintyawarelidarpanopticsegmentation} proposes a single-pass
uncertainty-aware LiDAR panoptic model on a polar BEV grid and reports evaluation results on SemanticKITTI of uncertainty-aware panoptic metrics and uncertainty calibration scores that are based on the Expected Calibration Error (ECE). 
ULOPS \cite{mohan2025opensetlidarpanopticsegmentation} extends Dirichlet-based uncertainty to open-set LiDAR panoptic segmentation in polar BEV, introducing uncertainty-driven losses for known/unknown separation, and compares multiple uncertainty estimation strategies on KITTI-360 \cite{Liao2022PAMI} and Panoptic nuScenes \cite{panoptic_nuScenes} that focuses primarily on open-set separation and accuracy-oriented outcomes rather than a systematic calibration study.

Overall, evidential LiDAR segmentation literature is concentrated in BEV and panoptic settings \cite{sirohi2022uncertaintyawarelidarpanopticsegmentation, mohan2025opensetlidarpanopticsegmentation}, where Dirichlet-family losses are typically adopted as fixed design choices rather than systematically analyzed. 
This leaves both their loss-design sensitivity and their extension to range-view semantic segmentation insufficiently studied. 
We address this gap through a controlled closed-set study of Dirichlet-family objectives for LiDAR range-view semantic segmentation under shared backbones and datasets, and introduce a Dirichlet-inspired parameterization that decouples class preference from evidence strength, calibrated by an inverse-vacuity self-calibration signal.
\section{\large Method}
\label{sec_method}
In this section, we present our evidential uncertainty modeling framework for LiDAR range-view semantic segmentation. 
We first introduce the Dirichlet formulation used in evidential deep learning (EDL) and the uncertainty measures used in this work.
In Sec.~\ref{sec:method_shapeLosses}, we summarize the Dirichlet-family training objectives used in prior LiDAR evidential segmentation work. 
Next, we present our \textit{Adapter Head} architecture (Sec.~\ref{sec:method_head}), which splits into a \textit{Preference Head} and a \textit{Strength Head} (see \autoref{fig:architectural_design}). 
From these two outputs, we construct data \textit{Evidence}, Dirichlet \textit{Concentration}, and the \textit{Dirichlet Probabilities} (Sec.~\ref{sec:method_reparam}). 
Finally, we introduce our \textbf{in}verse-\textbf{va}cuity \textbf{s}elf-\textbf{cal}ibration objective (Invascal) for training the \textit{Strength} signal (Sec.~\ref{sec:method_strength_loss}).

\subsection{Dirichlet preliminaries}
A Dirichlet distribution defines a distribution over categorical probability vectors $\vec p=(p_1,\dots,p_K)$ on the simplex
$\mathcal{S}=\{ \, \vec{p}:  p_k \ge 0, \; \sum_k p_k = 1 \, \}$ for $K$ classes. 
It is parameterized by \textit{Concentration} parameters $\vec{\alpha} = (\alpha_1,\dots,\alpha_K)$, and its probability density function (PDF) on vectors $\vec{p} \in \mathcal{S}$ is
\begin{equation}
\text{Dir}(\vec{p}\mid\vec{\alpha})
= \frac{1}{\mathrm{B}(\vec{\alpha})} \prod_{k=1}^K p_k^{\alpha_k-1},
\quad
\mathrm{B}(\vec{\alpha}) = \frac{\prod_{k=1}^K \Gamma(\alpha_k)}{\Gamma(\alpha_0)},
\end{equation}
with $\mathrm{B}(\cdot)$ the multivariate Beta function, $\Gamma(\cdot)$ is the Gamma function.
Intuitively, $\alpha_0=\sum_k \alpha_k$ controls how concentrated the distribution is.
Classical EDL parameterizes Dirichlet \textit{Concentration} via nonnegative class-wise \textit{Evidence} $e_k(\vec x)\ge 0$ and a base prior mass $b_k>0$,
\begin{equation}
\alpha_k(\vec x)=b_k+e_k(\vec x).
\label{eq:alpha_from_evidence}
\end{equation}
A common choice is a symmetric unit prior ($b_k=1$ for all $k$), although non-unit or learnable priors have also been studied \cite{chen2024revisitingessentialnonessentialsettings}.
The expected values of $\vec{p}$ (\textit{Dirichlet Probabilities}) are given by the Dirichlet mean
\begin{equation}
\mathbb{E}[p_k](\vec x)=\frac{\alpha_k(\vec x)}{\alpha_0(\vec x)}.
\label{eq:predictive_mean}
\end{equation}

\subsection{Uncertainty quantification}
In EDL, uncertainty is commonly quantified via the Dirichlet vacuity or uncertainty mass: 
\begin{equation}
u_{\text{EDL}}(\vec x)=\frac{\sum_{k=1}^K b_k}{\alpha_0(\vec x)}.
\label{eq:u_edl}
\end{equation}
Under the symmetric unit prior, this reduces to $u_{\text{EDL}}=K/\alpha_0$.
Thus, vacuity decreases as total \textit{Evidence} increases.

However, $u_{\text{EDL}}$ is only input-dependent on $\alpha_0$ and is therefore insensitive to how probability mass is distributed across classes.
For example, it cannot distinguish an ambiguous high-evidence case (e.g., $\vec\alpha=[50,50,1]$) from a confident unimodal one (e.g., $\vec\alpha=[100,1,1]$) if $\alpha_0$ is similar.
We therefore additionally use the normalized predictive entropy of the \textit{Dirichlet Probabilities} as in \cite{Tsiligkaridis_2021},
\begin{equation}
u_H(\vec x)=\frac{H_{\text{pred}}(\vec x)}{\log K},
\;
H_{\text{pred}}(\vec x)\!=\!
-\!\!\sum_{k=1}^K\! \mathbb{E}[p_k](\vec x) \log \mathbb{E}[p_k](\vec x),
\label{eq:dir-total-entropy}
\end{equation}
which is directly comparable to the Shannon entropy of \textit{Softmax Probabilities}.
Unlike vacuity, entropy is shape-sensitive: it increases when predictive mass is spread across classes and decreases when the distribution is concentrated.

\subsection{Dirichlet shape losses}
\label{sec:method_shapeLosses}
Given Dirichlet outputs $\vec\alpha(\vec x)$, 
evidential training typically minimizes a Dirichlet-family objective derived from a scoring rule $\ell(y,\vec p)$ for ground-truth class $y$:
\begin{equation}
\mathcal L(y\mid \vec\alpha)
=
\mathbb E_{\vec p\sim \mathrm{Dir}(\vec\alpha)}[\ell(y,\vec p)]
=
\int_{\mathcal S}\ell(y,\vec p)\,\mathrm{Dir}(\vec p\mid\vec\alpha)\,d\vec p.
\label{eq:dir-loss_formulation_general}
\end{equation}
For a broader discussion of evidential objectives, see \cite{juergens2024epistemicuncertaintyfaithfullyrepresented}. We consider three standard losses:
\begin{enumerate}[leftmargin=*]
\item \textit{Digamma (expected cross-entropy).}
Using $\ell(y,\vec p)=-\log p_y$ yields
\begin{equation}
\mathcal L_{\text{Digamma}}(y\mid\vec\alpha)
=
\mathbb E_{\vec p\sim \mathrm{Dir}(\vec\alpha)}[-\log p_y]
=
\psi(\alpha_0)-\psi(\alpha_y),
\label{eq:dir-digamma-ce}
\end{equation}
where $\psi(\cdot)$ is the digamma function.

\item \textit{NLL (negative log-marginal likelihood).}
Using the predictive mean as \textit{Dirichlet Probabilities} gives
\begin{equation}
\begin{split}
\mathcal L_{\text{NLL}}(y \mid \vec\alpha)
&= -\log p(y \mid \vec\alpha) = -\log \mathbb E[p_y] \\
&= \log\alpha_0-\log\alpha_y.
\end{split}
\label{eq:dir-nll}
\end{equation}

\item \textit{MSE (expected Brier score).}
Let $\delta_{y,k}\in\{0,1\}$ be the one-hot target, then
\begin{equation}
\mathcal L_{\text{MSE}}(y\mid\vec\alpha)
=
\sum_{k=1}^K
\Big[
\big(\delta_{y,k}-\mathbb E[p_k]\big)^2 + \mathrm{Var}[p_k]
\Big].
\label{eq:dir-mse}
\end{equation}
\end{enumerate}
Prior work reports that Dirichlet objectives can produce inflated Dirichlet \textit{Concentration} and overconfident \textit{Evidence} \cite{sensoy2018evidentialdeeplearningquantify,Tsiligkaridis_2021,charpentier2020posteriornetworkuncertaintyestimation,chen2024revisitingessentialnonessentialsettings}.
A common remedy is to add a Kullback-Leibler (KL) regularizer $\mathcal{R}_{\mathrm{KL}}(y\mid\vec\alpha)$ that suppresses spurious off-class \textit{Evidence} by pulling non-ground-truth classes toward the base prior.

Among existing LiDAR evidential segmentation methods, 
EvLPSNet \cite{sirohi2022uncertaintyawarelidarpanopticsegmentation} uses NLL+KL, 
ULOPS \cite{mohan2025opensetlidarpanopticsegmentation} uses Digamma, 
and the original EDL formulation \cite{sensoy2018evidentialdeeplearningquantify} uses MSE+KL.
In contrast to these Dirichlet PDF shaping objectives, which couple class \textit{Preference} and \textit{Concentration} through the same per-class parameters and often rely on KL to indirectly control \textit{Evidence}, our method explicitly separates \textit{Preference} (class ranking) from \textit{Strength} (total concentration) and calibrates \textit{Strength} with a dedicated inverse-vacuity objective.

\begin{figure*}
    \centering
    \includegraphics[width=0.9\linewidth,trim=0cm 0cm 0cm 0cm,clip]
    {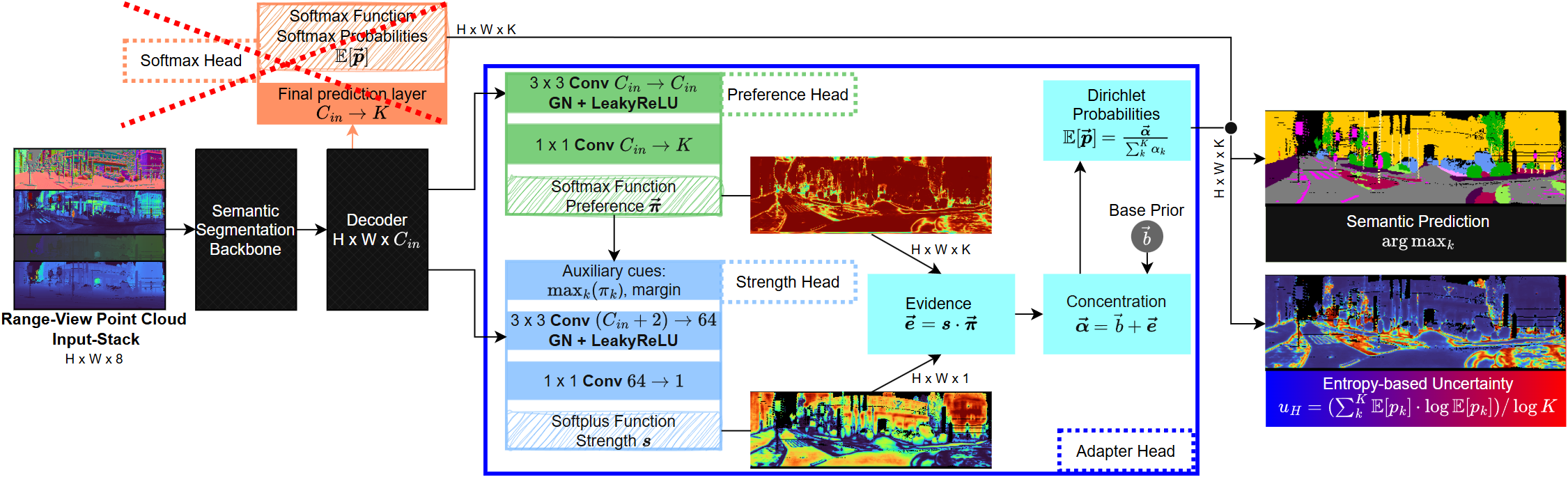}
    \vspace{-0.1cm}
    \caption{Overview of our Invascal architecture.}
    \label{fig:architectural_design}
\end{figure*}
\subsection{Adapter Head: Preference Head and Strength Head}
\label{sec:method_head}
An overview of our architectural design is shown in \autoref{fig:architectural_design}.
We attach an \textit{Adapter Head} to the decoder output
$
f(\vec x)\in\mathbb R^{H\times W\times C_{\mathrm{in}}},
$
i.e., the final feature map that would otherwise be passed directly to the backbone's original \textit{Final prediction layer} ($C_{\mathrm{in}}\!\rightarrow\!K$) and softmax activation function. 
The backbone and decoder up to $f(\vec x)$ remain unchanged.
Our adapter replaces this final mapping with two parallel branches at the same spatial resolution $H\times W$.

The \textit{Preference Head} produces class logits $\vec z(\vec x)\in\mathbb{R}^{K}$ and the corresponding \textit{Preference} distribution 
$\vec\pi(\vec x)=\mathrm{softmax}(\vec z(\vec x))$.
Starting from $f(\vec x)$, the first $3{\times}3$ convolution preserves channel width and spatial resolution 
and acts as a local refinement before collapsing the representation to class scores.

The \textit{Strength Head} predicts a scalar \textit{Strength} map $s(\vec x) \ge 0$, which controls \textit{Evidence} magnitude (certainty) independently of class ranking.
It operates on the richer decoder features $f(\vec x)$ rather than only on the $K$ logits, because the logits are already a compressed semantic representation and may discard cues relevant for \textit{Evidence} support.
Optionally, we concatenate two auxiliary confidence cues to $f(\vec x)$ derived from $\vec\pi(\vec x)$:
(i) the maximum \textit{Preference} probability $\max_k \pi_k(\vec x)$ and
(ii) the top-1/top-2 margin.
A softplus activation enforces nonnegative \textit{Strength} $s(\vec x)$.

We optionally detach $f(\vec x)$ on the \textit{Strength Head} path so that our Invascal loss (see Sec.~\ref{sec:method_strength_loss}) does not alter the backbone features used by the \textit{Preference Head}. 
The auxiliary cues are computed from $\vec\pi(\vec x)$ without gradient propagation into the \textit{Preference Head}. In our experiments, this stabilizes \textit{Preference} learning and improves entropy-based calibration.

\subsection{From Preference and Strength to Dirichlet Evidence and Concentration}
\label{sec:method_reparam}
In classical EDL, the same per-class outputs jointly determine both class preference (via $\mathbb E[p_k](\vec x)$ from Eq.~\eqref{eq:predictive_mean}) and certainty (via the total \textit{Concentration} $\alpha_0$), which couples semantic ranking and certainty level. 
We decouple these roles by the \textit{Preference} $\vec\pi(\vec x)$ and scalar \textit{Strength} $s(\vec x)$ produced by the \textit{Adapter Head}.
We construct class-wise data \textit{Evidence} 
\begin{equation}
e_k(\vec x)=s(\vec x)\,\pi_k(\vec x),
\label{eq:evidence_from_strength}
\end{equation}
which 
preserves the ordering induced by $\vec\pi$ and lets a single scalar control total \textit{Evidence} without introducing additional class-wise certainty parameters. 
Hence, the model can attenuate or amplify confidence without changing the preferred class.
\textit{Preference} training is also loss-agnostic, allowing us to adopt any state-of-the-art loss trained on \textit{Softmax Probabilities}.

With Eq.~\eqref{eq:evidence_from_strength}, the total \textit{Evidence} is
$\sum_{k=1}^K e_k(\vec x)=s(\vec x).$
Using the standard EDL relation in Eq.~\eqref{eq:alpha_from_evidence}, the Dirichlet \textit{Concentration} become
\begin{equation}
\alpha_k(\vec x)\!=\!b_k+e_k(\vec x)\!=\!b_k+s(\vec x)\pi_k(\vec x),
\quad
\alpha_0(\vec x)\!=\!\sum_{k=1}^K b_k+s(\vec x).
\label{eq:alpha_reparam}
\end{equation}
The resulting \textit{Dirichlet Probabilities} are
\begin{equation}
\mathbb E[p_k](\vec x)
=
\frac{b_k+s(\vec x)\pi_k(\vec x)}{\sum_{j=1}^K b_j+s(\vec x)}.
\label{eq:mean_reparam}
\end{equation}
For a symmetric prior $b_k=b$, this becomes
\begin{equation}
\mathbb E[p_k](\vec x)
=
\underbrace{\frac{Kb}{Kb+s(\vec x)}}_{\text{prior weight}}\,\frac{1}{K}
+
\underbrace{\frac{s(\vec x)}{Kb+s(\vec x)}}_{\text{evidence weight $q(\vec{x})$}} \!\!\pi_k(\vec x),
\label{eq:predictive_mean_uniform_prior_reparam}
\end{equation}
i.e., a convex combination of the uniform prior and the \textit{Preference} distribution. 
This makes the decoupling explicit: $s(\vec x)$ changes certainty through $\alpha_0(\vec x)$ while preserving the class order induced by $\vec\pi(\vec x)$.

In the architecture diagram, the final semantic prediction is the argmax of the \textit{Dirichlet Probabilities}
and entropy-based uncertainty $u_H$ is computed from the same predictive mean using Eq.~\eqref{eq:dir-total-entropy}.

\paragraph*{Example}
Consider $K=3$ with symmetric unit prior $b=1$. If the \textit{Preference Head} predicts
$\vec\pi=[0.90,0.09,0.01]$ but the \textit{Strength Head} outputs low \textit{Strength} $s=1$, then
$\mathbb{E}[\vec p\,] = \frac{1}{3+1}\,[1,1,1] + \frac{1}{3+1}\,[0.90,0.09,0.01]
\approx [0.48,0.27,0.25]$,
so the model remains cautious despite a peaky \textit{Preference}. 
If instead $s=30$, then
$\mathbb{E}[\vec p\,] \approx \frac{3}{33}\frac{1}{3} + \frac{30}{33}\vec\pi \approx [0.85,0.10,0.05]$,
which becomes close to $\vec\pi$, reflecting high \textit{Evidence} support for the same class ranking.

\subsection{Strength calibration objective (Invascal)}
\label{sec:method_strength_loss}
Under Eq.~\eqref{eq:predictive_mean_uniform_prior_reparam}, the contribution of the \textit{Preference} distribution is governed by the \textit{Evidence} weight
\begin{equation}
q(\vec x)=\frac{s(\vec x)}{Kb+s(\vec x)} = 1-u_{\mathrm{EDL}}(\vec x)\in (0,1) ,
\label{eq:q_def}
\end{equation}
which is the complement of Dirichlet vacuity from Eq.~\eqref{eq:u_edl}. 
Learning $s(\vec x)$ is therefore equivalent to learning an input-dependent inverse-vacuity signal.

We supervise this signal using a self-calibration target derived from the \textit{Preference Head}. Let $y(\vec x)$ be the ground-truth class and define
$p_y(\vec x)=\pi_{y(\vec x)}(\vec x)$,
i.e., the \textit{Preference} probability assigned to the correct class. 
We stop gradients through $p_y(\vec x)$ so that this objective updates only the \textit{Strength Head}.

To avoid saturating the \textit{Strength} target on easy pixels, we impose a minimum uncertainty floor $u_{\min}\in(0,1)$ and cap the target:
\begin{equation}
q_{\max}=1-u_{\min},
\quad
c(\vec x)=\min\{\mathrm{stopgrad}(p_y(\vec x)),\,q_{\max}\}.
\label{eq:c_target}
\end{equation}
This cap prevents the calibration loss from encouraging arbitrarily large \textit{Concentration} in regions where the \textit{Preference Head} is already highly confident.

We then define the Invascal loss as a point-wise binary cross-entropy between predicted inverse-vacuity $q(\vec x)$ and capped target $c(\vec x)$, averaged over valid pixels $\Omega$:
\begin{equation}
\mathcal L_{\text{Invascal}}\!=\!-\frac{1}{|\Omega|}\!\sum_{\vec x\in\Omega}\!\big[c(\vec x)\log q(\vec x)\!+\!(1\!-\!c(\vec x))\log(1\!-\!q(\vec x))\big].
\label{eq:strength_bce}
\end{equation}
This objective increases \textit{Strength} where the \textit{Preference Head} assigns strong support to the correct class and decreases \textit{Strength} where that support is weak. 
Through Eq.~\eqref{eq:predictive_mean_uniform_prior_reparam}, it attenuates overconfident predictions in low-evidence regions while preserving the semantic ranking encoded by $\vec\pi(\vec x)$.
In training, we combine $\mathcal L_{\text{Invascal}}$ with the standard segmentation loss $\mathcal L_{\text{DET}}$ applied to the \textit{Preference Head} output $\vec\pi(\vec x)$ through linear combination.
\section{\large Experiments}
\label{sec:expirement}

\begin{figure*}[!htbp]
    \centering
    \begin{subfigure}[c]{0.4999\textwidth}
        \includegraphics[width=\linewidth,trim=0 0.3cm 0 0.1,clip]
        {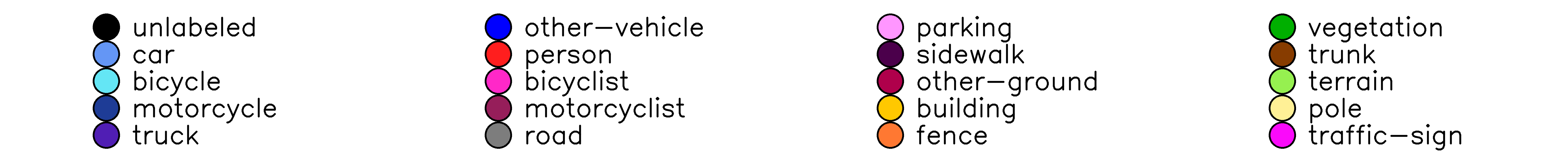}
        \vspace{-1.8em}
        \caption*{Legend of semantic labels. Mapping of color to class name.}
    \end{subfigure}\hfill
    \begin{subfigure}[c]{0.4999\textwidth}
        \includegraphics[width=\linewidth,trim=0 0.3cm 0 0.1,clip]
        {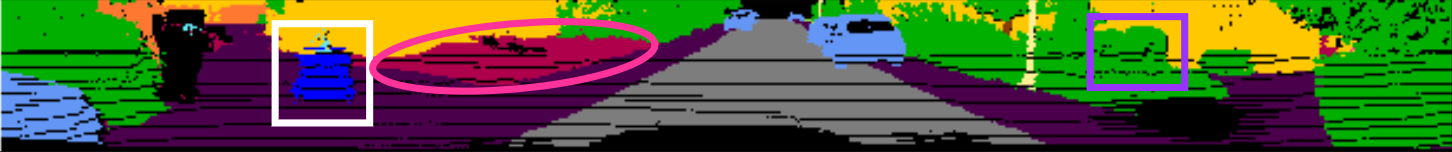}
        \vspace{-1.8em}
        \caption*{Ground truth labels}
    \end{subfigure}
    \vspace{0.1cm}
    \rule{\linewidth}{0.4pt}
    \vspace{-0.4cm}

    \begin{subfigure}[c]{\textwidth}
        \includegraphics[width=\linewidth,trim=0 0.3cm 0 0.1,clip]
        {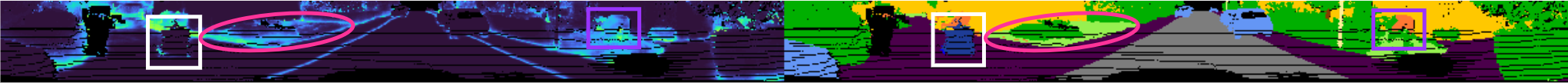}
        \vspace{-1.8em}
        \caption{DET}
    \end{subfigure}\hfill 
    
    \begin{subfigure}[c]{\textwidth}
        \includegraphics[width=\linewidth,trim=0 0.3cm 0 0.1,clip]
        {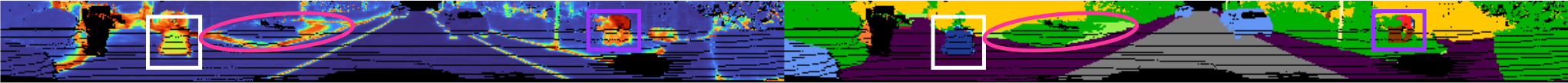}
        \vspace{-1.8em}
        \caption{NLL+KL}
    \end{subfigure}\hfill
    
    \begin{subfigure}[c]{\textwidth}
        \includegraphics[width=\linewidth,trim=0 0.3cm 0 0.1,clip]
        {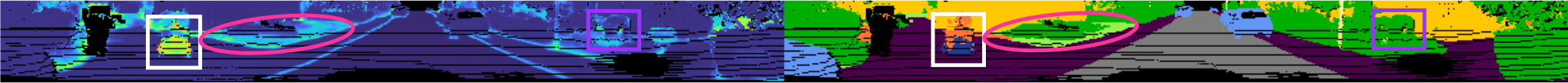}
        \vspace{-1.8em}
        \caption{Digamma}
    \end{subfigure}\hfill
    
    \begin{subfigure}[c]{\textwidth}
        \includegraphics[width=\linewidth,trim=0 0.3cm 0 0.1,clip]
        {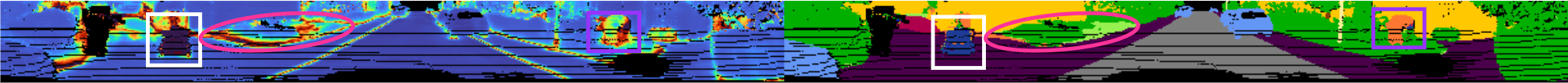}
        \vspace{-1.8em}
        \caption{MSE+KL}
    \end{subfigure}\hfill
    
    \begin{subfigure}[c]{\textwidth}
        \includegraphics[width=\linewidth,trim=0 0.3cm 0 0.1,clip]
        {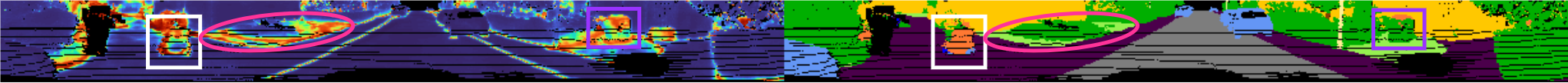}
        \vspace{-1.8em}
        \caption{Invascal}
    \end{subfigure}\hfill  

    \vspace{-0.8em}
    \caption{\small
        Qualitative results on SemanticKITTI validation set trained on SalsaNext model with different loss functions. The entropy-based uncertainty images $u_{H}$ are on the left side. Low uncertainty is shown in blue, moderate in green to yellow, and high in red (i.e., Turbo Colormap \cite{google2019turbocolormap}). On the right, the semantic predictions are shown. 
        Note that this example is not accuracy-focused, and rather focuses in how uncertainty is expressed when the model is wrong.
    }
    \label{fig:qualitative_results}
\end{figure*}


\subsection{Datasets}\label{sec:datasets}
We evaluate on three public LiDAR datasets covering diverse environments and sensor setups:
SemanticKITTI \cite{SemanticKITTI_behley2019iccv} (suburban),
SemanticTHAB \cite{SemanticTHAB_reichert_2025_14906179} (urban),
and Panoptic-CUDAL \cite{tseng2025panopticcudaltechnicalreportrural} (rural and rainy).
Together, they span different traffic densities, weather conditions, and LiDAR configurations (64--128 channels with varying fields of view), enabling evaluation under domain and condition shifts.

All methods operate on LiDAR range-view projections \cite{reichert2023sensorequivariancelidarprojection}.
We use image sizes $64{\times}1024$ (SemanticKITTI), $128{\times}1280$ (SemanticTHAB), and $128{\times}1024$ (Panoptic-CUDAL).
Each pixel contains Euclidean range, Cartesian coordinates $(x,y,z)$, Reflectivity, and 3D surface normals.


\subsection{Experimental Setup}
For our experiments, we use two LiDAR range-view semantic segmentation baseline models to prove consistency of our method:
(i) SalsaNext \cite{SalsaNext}, which includes dropout in encoder/decoder blocks and is therefore compatible with Bayesian approximations, and
(ii) the model of Reichert et al. \cite{reichert2025realtimesemanticsegmentation}, which is equipped with an EfficientNetV2-L backbone \cite{tan2021efficientnetv2smallermodelsfaster}.

For each backbone, we train five variants:
(a) \textit{DET}: cross-entropy + Lov\'asz-Softmax \cite{SalsaNext} producing \textit{Softmax Probabilities} (determinstic softmax baseline).
(b) \textit{NLL+KL} \cite{sirohi2022uncertaintyawarelidarpanopticsegmentation}, (c) \textit{Digamma} \cite{mohan2025opensetlidarpanopticsegmentation}, and (d) \textit{MSE+KL} \cite{sensoy2018evidentialdeeplearningquantify}, that are all Dirichlet-family losses from Sec.~\ref{sec:method_shapeLosses}.
(e) \textit{Invascal}, our proposed method with the \textit{Adapter Head} (Sec.~\ref{sec:method_head}) and the decoupled \textit{Preference} and \textit{Strength} learning (Sec.~\ref{sec:method_strength_loss}).
For KL-based losses, we follow the KL weight scheduling of EvLPSNet \cite{sirohi2022uncertaintyawarelidarpanopticsegmentation}.
For Invascal, we use linear combination weights $\lambda_{\text{Invascal}}=0.2$ and $\lambda_{\text{DET}}=1$, and we set $u_{\min}=0.01$ in Eq.~\eqref{eq:c_target}.

All models are trained for 40 epochs with peak learning rate $5{\times}10^{-4}$, using linear warm-up from 0 to its peak for 4 epochs and cosine decay to $5{\times}10^{-7}$, following \cite{ando2023rangevitvisiontransformers3d}.
Augmentations are horizontal flip in range-view (azimuth reversal) with probability $0.5$ and yaw rotation in $\pm 30^\circ$.
We use fixed random seeds and deterministic training settings to reduce variance from initialization, data loading, augmentation, and dropout.
Unlabeled or void pixels are ignored during both training and evaluation.

\subsection{Qualitative Results}
\autoref{fig:qualitative_results} compares semantic predictions and entropy-based uncertainty maps to assess how well uncertainty aligns with errors.
We focus on three highlighted regions with coherent mistakes across methods: the white box (class \textit{other-vehicle}), the pink ellipse (class \textit{other-ground}), and the purple box (class \textit{vegetation}).
The first two are inherently difficult due to their generic category, 
e.g., \textit{other-vehicle} could not be assigned to any of the vehicular type classes car, truck, motorcycle, bicycle.
The third example reflects a common confusion between visually similar classes.

In the white box, none of the methods predict the ground truth class other-ground. 
MSE+KL and DET stand out because they assign low uncertainty to an incorrect motorcycle prediction, indicating overconfident errors.
The other methods place noticeably higher uncertainty on the object, with NLL+KL producing the most spatially consistent uncertain region.
Invascal mostly predicts the class fence (orange), with other class predictions mixed into the objects' contours, where the model correctly assigns higher uncertainty.
In the pink ellipse, Digamma and DET appear most confident despite the error, whereas the others show stronger attenuation.
In the purple box, only Digamma correctly predicts the class.
Among the incorrect predictions, several methods confuse the region with a fence (orange). 
NLL+KL also predicts trunk (brown) and person (red), with the person pixels marked with the highest uncertainty. The wrong brown pixels and the correct vegetation have the same predictive ambiguity.
Our method localizes higher uncertainty clearly around the ambiguous boundary, while DET shows a broader, less informative uncertainty pattern, and MSE+KL concentrates uncertainty less consistently with the error region.

Overall, MSE+KL exhibits the most problematic overconfident failures in this example, while DET and Digamma often yield visually sharp predictions with limited uncertainty contrast.
The strongest qualitative alignment between errors and uncertainty is observed for NLL+KL and Invascal.
We next quantify these trends using segmentation performance and calibration metrics.

\subsection{Segmentation Performance}
\autoref{table:miouresults} reports class-wise IoU, mIoU, and calibration metrics across all datasets and backbones.
Across all datasets and backbones, Invascal differs from DET by +0.1 mIoU points on average, indicating that the proposed Invascal loss is effectively mIoU-neutral and does not introduce systematic accuracy degradation.
In contrast, the Dirichlet variants can substantially reduce segmentation performance. 
Most notably for SemanticKITTI/SalsaNext, where performance drops sharply on safety-relevant classes such as Person and Bicycle.
In particular, MSE+KL decreases mIoU by -13.2 pp relative to DET, which is the largest deviation observed across all settings.

Averaged over all settings, Invascal achieves the highest mean mIoU (53.45\%), closely followed by DET (53.35\%), then NLL+KL (52.13\%) and Digamma (51.48\%), while MSE+KL yields the lowest mean mIoU (49.02\%). 
These results highlight the robustness of Invascal across datasets and backbones, while evidential Dirichlet variants can incur large accuracy regressions in certain regimes.

\begin{table*}[!htbp]
\centering
\caption{\small Calibration comparison on Panoptic-CUDAL using SalsaNext, focusing on deterministic calibration baselines versus Invascal. We compare the deterministic baseline (DET), post-hoc Temperature Scaling applied to DET (DET+TS) \cite{guo2017calibrationmodernneuralnetworks}, and MC Dropout with 50 stochastic forward passes (DET+MC), against Invascal. Temperature scaling is fitted on 25\% of the Panoptic-CUDAL validation set by minimizing negative log-likelihood, and all methods are evaluated on the same remaining 75\% split. We report per-class ECE and entropy-based uECE, as well as aggregate ECE and uECE, all in percent. Inference speed is reported in frames per second (FPS) on a NVIDIA RTX 4090.
}
\label{table:abl}
\vspace{-0.2em} 
\scriptsize
\setlength{\tabcolsep}{4pt}
\renewcommand{\arraystretch}{0.95}
\resizebox{0.99\textwidth}{!}{
\begin{tabular}{|c   *{9}{|c} |c  c c |}
\hline
Loss & \multicolumn{9}{c|}{per class - ECE | uECE} & ECE & uECE 
& Speed(FPS) \\
\cline{2-10}
& \rotatebox[origin=c]{0}{Person}
& \rotatebox[origin=c]{0}{Car}
& \rotatebox[origin=c]{0}{Road}
& \rotatebox[origin=c]{0}{Sidewalk}
& \rotatebox[origin=c]{0}{Building}
& \rotatebox[origin=c]{0}{Vegetation}
& \rotatebox[origin=c]{0}{Terrain}
& \rotatebox[origin=c]{0}{Pole}
& \rotatebox[origin=c]{0}{Traffic-sign}
&&&
\\

\hline
DET & 62.9 \,|\, 61.6 & 4.6 \,|\, 4.7 & 4.9 \,|\, 5.1 & 40.7 \,|\, 41.4 & 6.9 \,|\, 7.0 & 3.0 \,|\, 3.0 & 0.7 \,|\, 0.7 & 8.1 \,|\, 8.2 & 4.9 \,|\, 5.1 & 4.8 & 4.9 
& 70.8 \\

DET+TS & 57.5 \,|\, 54.5 & 4.2 \,|\, 4.1 & 3.9 \,|\, 3.8 & 37.6 \,|\, 37.8 & 5.5 \,|\, 5.1 & 1.9 \,|\, \textbf{1.4} & \textbf{0.2} \,|\, 1.1 & 7.0 \,|\, 6.8 & 3.5 \,|\, \textbf{3.3} & 3.7 & 3.3 
& \textbf{71.3} \\

DET+MC & 54.5 \,|\, 53.9 & \textbf{1.2} \,|\, \textbf{1.9} & 4.1 \,|\, 4.4 & \textbf{35.1} \,|\, 37.3 & 4.6 \,|\, 5.0 & 2.3 \,|\, 2.4 & 0.3 \,|\, \textbf{0.3} & \textbf{5.6} \,|\, 6.1 & 3.0 \,|\, \textbf{3.3} & 3.9 & 4.1 
& 3.6 \\

Invascal & \textbf{54.2} \,|\, \textbf{44.3} & 1.3 \,|\, 2.7 & \textbf{3.4} \,|\, \textbf{1.7} & 43.6 \,|\, \textbf{36.6} & \textbf{2.4} \,|\, \textbf{2.9} & \textbf{1.2} \,|\, 2.9 & 1.5 \,|\, 4.5 & 6.9 \,|\, \textbf{3.9} & \textbf{2.1} \,|\, 4.2 & \textbf{2.8} & \textbf{1.7} 
& 64.9 \\


\hline
\end{tabular}
}
\renewcommand{\arraystretch}{1.0}
\setlength{\tabcolsep}{6pt}
\end{table*}

\begin{table}[!htbp]
\centering
\caption{\small Per class IoU (selected classes) and mIoU, accompanied by calibration-focused metrics: ECE and uncertainty ECE using entropy as inverse confidence (uECE). All scores are reported in percent. For the mean scores, we color the best result per configuration in red and the second in blue. Classes not present in the test sets are denoted with $\, \text{\xmark} \,$ and are not considered in the metrics.}
\label{table:miouresults}
\vspace{-0.8em} 
\scriptsize
\setlength{\tabcolsep}{4pt} 
\renewcommand{\arraystretch}{0.98}   
\resizebox{\columnwidth}{!}{%
\begin{tabular}{| c | c | c | *{4}{c} | c c c |}
\hline
Dataset & Model & Loss & \multicolumn{4}{c|}{IoU of Road Users} & mIoU & ECE & uECE 
\\
\cline{4-7}
&&
& \rotatebox[origin=c]{90}{Person}
& \rotatebox[origin=c]{90}{Bicycle}
& \rotatebox[origin=c]{90}{Car}
& \rotatebox[origin=c]{90}{Truck}
&&&
\\
\hline

\multirow{10}{*}{\rotatebox[origin=c]{90}{SemanticKITTI}}
& \multirow{5}{*}{\rotatebox[origin=c]{90}{SalsaNext}}
& DET & 63.7 & 32.2 & 95.5 
& 81.2 
& \textcolor{red}{59.7}
& 6.0
& 6.0
\\
& & NLL+KL & 36.0 & 2.1 & 95.8 
& 62.7 
& 50.6
& \textcolor{blue}{3.3}
& 4.6
\\
& & Digamma & 39.9 & 0.0 & 95.4 
& 0.0 
& 47.3 
& 3.7
& \textcolor{red}{1.6}
\\
& & MSE+KL & 3.7 & 0.0 & 95.4 
& 60.9 
& 46.5 
& \textcolor{red}{2.5}
& 10.5
\\
& & Invascal & 63.5 & 41.1 & 95.2 
& 75.6 
& \textcolor{blue}{59.0}
& 4.6
& \textcolor{blue}{4.1}
\\

\cline{2-10}
& \multirow{5}{*}{\rotatebox[origin=c]{90}{Reichert}}
& DET & 53.1 & 16.4 & 93.9 
& 47.2 
& 52.9 
& 6.3
& 6.6
\\
& & NLL+KL & 48.8 & 21.1 & 95.0 
& 63.0 
& \textcolor{red}{55.5} 
& \textcolor{blue}{3.3}
& \textcolor{blue}{3.1}
\\
& & Digamma & 52.3 & 16.3 & 94.3 
& 66.4 
& \textcolor{blue}{54.9}
& 4.7
& 4.1
\\
& & MSE+KL & 7.7 & 0.0 & 95.3 
& 60.8 
& 46.9 
& \textcolor{red}{1.3}
& 7.1
\\
& & Invascal & 53.0 & 20.0 & 94.6 
& 74.5 
& 54.8 
& 4.1
& \textcolor{red}{2.9}
\\


\hline
\multirow{10}{*}{\rotatebox[origin=c]{90}{SemanticTHAB}}
& \multirow{5}{*}{\rotatebox[origin=c]{90}{SalsaNext}}
& DET & 72.6 & \xmark & 92.5 
& 77.3 
& \textcolor{red}{49.2} 
& 9.5
& 9.4
\\
& & NLL+KL & 66.8 & \xmark & 92.4 
& 77.6 
& 47.9 
& \textcolor{red}{3.4}
& 6.2
\\
& & Digamma & 68.4 & \xmark & 92.3 
& 78.7 
& 48.2 
& 5.1
& \textcolor{blue}{5.0}
\\
& & MSE+KL & 13.3 & \xmark & 92.2 
& 76.0 
& 44.0 
& \textcolor{blue}{4.6}
& 11.2
\\
& & Invascal & 73.9 & \xmark & 92.8 
& 75.5 
& \textcolor{blue}{49.1} 
& 5.9
& \textcolor{red}{3.7}
\\

\cline{2-10}
& \multirow{5}{*}{\rotatebox[origin=c]{90}{Reichert}}
& DET & 71.4 & \xmark & 85.5 
& 9.4 
& 45.1 
& 10.6
& 10.9
\\
& & NLL+KL & 70.3 & \xmark & 86.6 
& 18.5 
& \textcolor{blue}{46.1}
& \textcolor{blue}{6.0}
& \textcolor{red}{3.4}
\\
& & Digamma & 71.9 & \xmark & 86.3 
& 22.6 
& \textcolor{red}{46.6}
& 8.1
& 6.9
\\
& & MSE+KL & 70.5 & \xmark & 85.7 
& 12.3 
& 45.3
& \textcolor{red}{3.8}
& \textcolor{blue}{4.9}
\\
& & Invascal & 75.4 & \xmark & 86.3 
& 12.0 
& \textcolor{blue}{46.1} 
& 7.3
& \textcolor{blue}{4.9}
\\


\hline
\multirow{10}{*}{\rotatebox[origin=c]{90}{Panoptic-Cudal}}
& \multirow{5}{*}{\rotatebox[origin=c]{90}{SalsaNext}}
& DET & 21.0 & \xmark & 85.5 
& 10.7 
& \textcolor{red}{57.9} 
& 4.8
& 4.9
\\
& & NLL+KL & 0.0 & \xmark & 86.7 
& 3.1 
& 56.4 
& \textcolor{blue}{2.4}
& 2.9
\\
& & Digamma & 0.0 & \xmark & 84.8 
& 6.9 
& 56.3 
& 3.7
& \textcolor{blue}{2.4}
\\
& & MSE+KL & 0.0 & \xmark & 86.7 
& 7.7 
& 55.7 
& \textcolor{red}{1.5}
& 6.1
\\
& & Invascal & 22.8 & \xmark & 87.7 
& 6.7 
& \textcolor{blue}{56.6} 
& 2.7
& \textcolor{red}{1.7}
\\

\cline{2-10}
& \multirow{5}{*}{\rotatebox[origin=c]{90}{Reichert}}
& DET & 0.3 & \xmark & 87.9 
& 2.8 
& 55.3 
& 4.8
& 4.9
\\
& & NLL+KL & 0.0 & \xmark & 88.0 
& 8.5 
& \textcolor{red}{56.3}
& \textcolor{blue}{2.4}
& 2.9
\\
& & Digamma & 0.0 & \xmark & 88.3 
& 5.4 
& 55.6 
& 3.7
& \textcolor{blue}{2.4}
\\
& & MSE+KL & 0.0 & \xmark & 88.1 
& 8.7
& \textcolor{blue}{55.7}
& \textcolor{red}{1.5}
& 6.1
\\
& & Invascal & 0.0 & \xmark & 88.3 
& 7.4
& 55.1 
& 3.6
& \textcolor{red}{1.7}
\\
\hline
\end{tabular}
}
\renewcommand{\arraystretch}{1.0}
\setlength{\tabcolsep}{6pt}
\end{table}

\subsection{Confidence and Uncertainty Calibration Analysis}
\begin{figure}[!htbp]
    \centering

    \includegraphics[width=0.8\columnwidth]{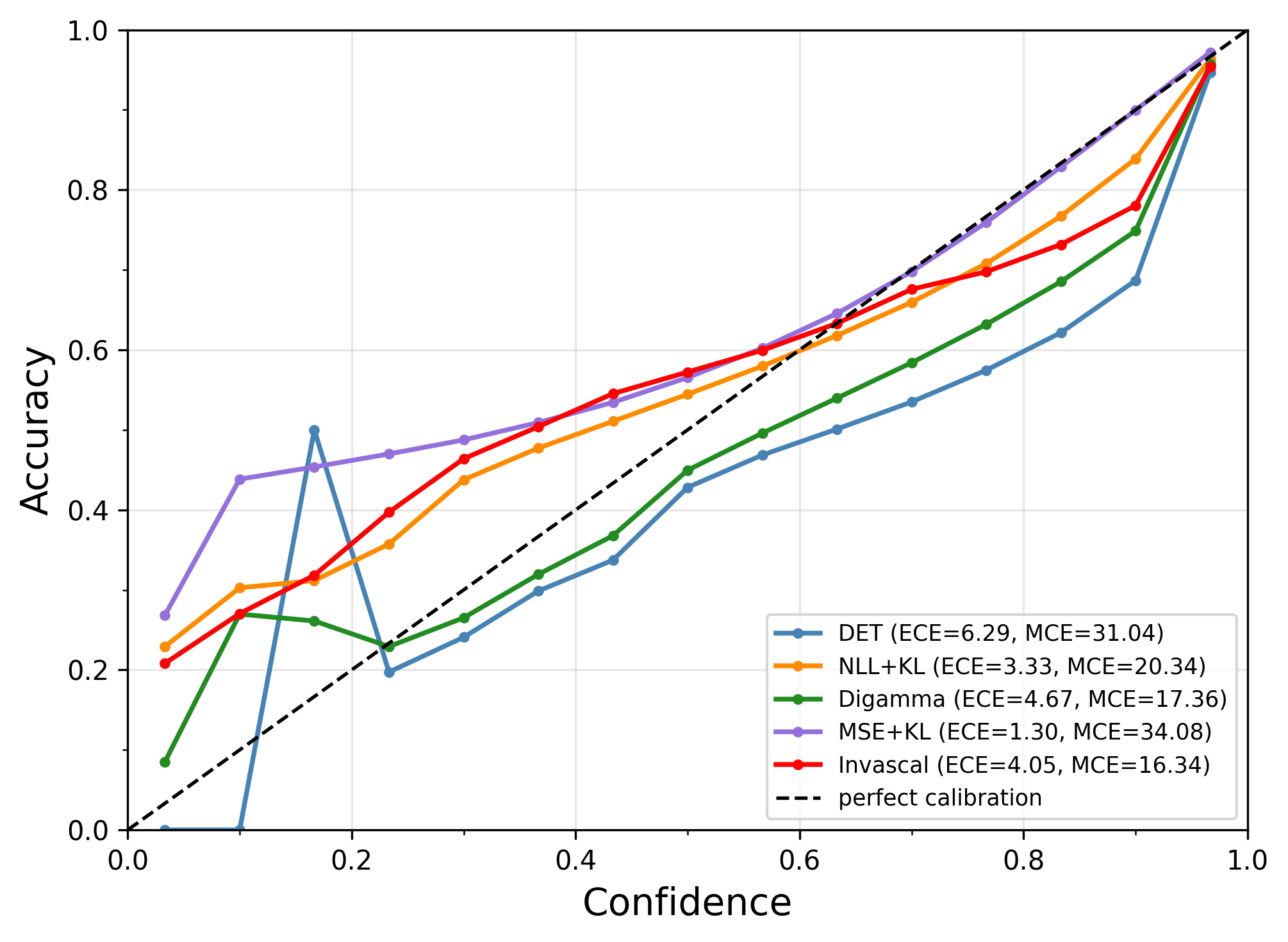}\\[-0.5em]
    \caption*{\small (a) Reliability diagrams using predicted expected probabilities as confidence. Lower ECE and MCE are better. Scores are reported in percent.
    }
    
    \vspace{0.6em}

    \includegraphics[width=0.8\columnwidth]{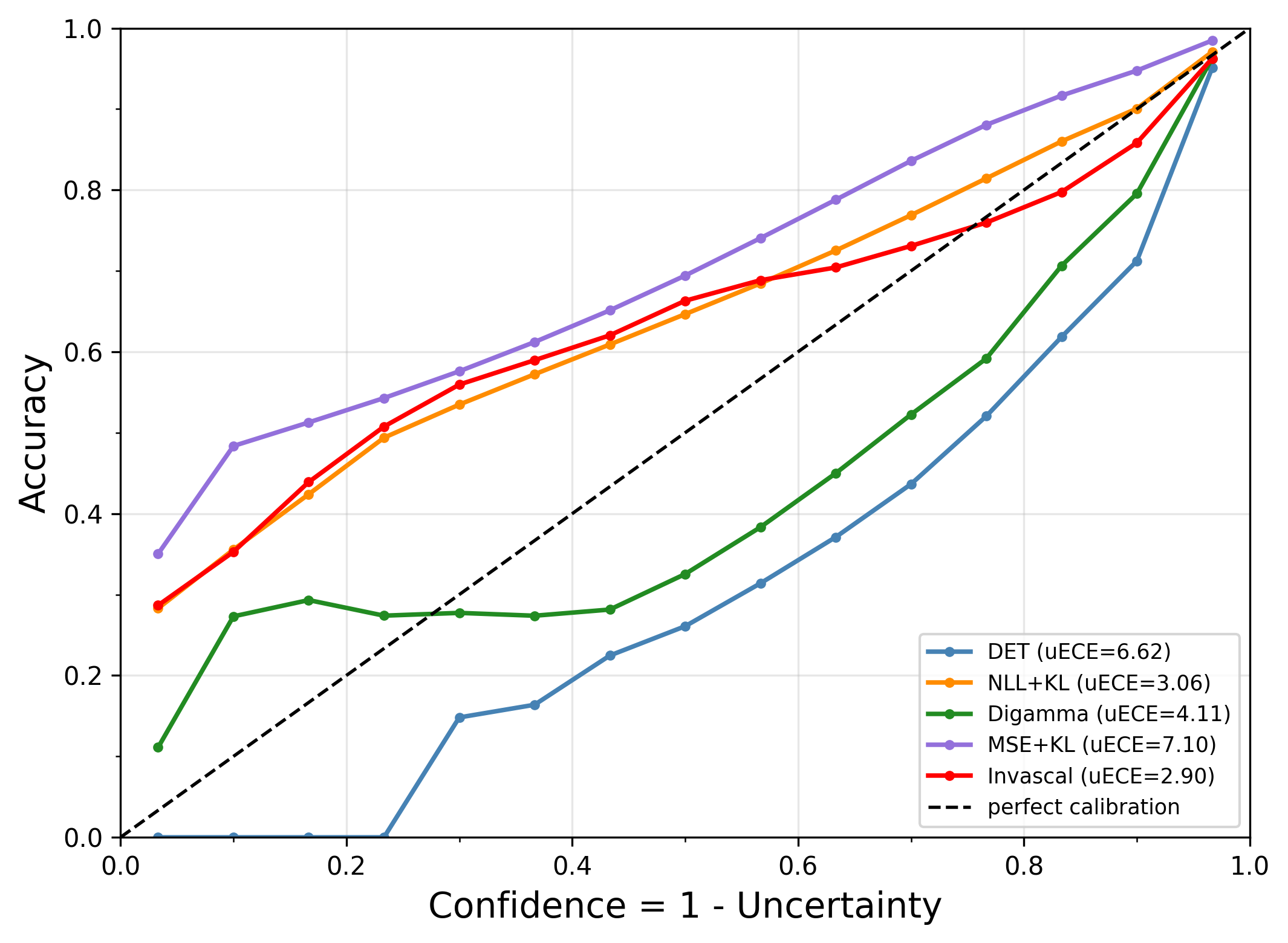}\\[-0.5em]
    \caption*{\small (b) Reliability diagrams using inverse uncertainty $1-u_H$ as confidence proxy. Lower uECE is better. Scores are reported in percent.}
    \caption{\small Calibration analysis of predicted probabilities and predicted uncertainty for different loss functions on the SemanticKITTI validation set using the model of \cite{reichert2025realtimesemanticsegmentation}.
    }
    \label{fig:reliability-diagrams-conf-unc}
\end{figure}

We assess calibration from two complementary viewpoints: calibration of the predicted class probabilities and calibration of the predicted uncertainty as a reliability signal.

\paragraph{Calibration of predicted probabilities}
For each valid pixel, we use the expected class probabilities as the predictive distribution and define confidence as the maximum expected probability,
$\text{conf}_p=\max_k \mathbb{E}[p_k]$.
\autoref{fig:reliability-diagrams-conf-unc}(a) reports reliability diagrams together with Expected Calibration Error (ECE) and Maximum Calibration Error (MCE) \cite{guo2017calibrationmodernneuralnetworks}.

The deterministic baseline is clearly overconfident and places few samples in low-confidence bins.
Among the Dirichlet objectives, adding the KL term generally reduces average miscalibration, as seen in \autoref{table:ECE_Dirichlet_w_wo_KL}, but this does not necessarily improve uniform calibration across the confidence range.
This is most visible for MSE+KL, which attains a low ECE but shows larger local deviations, reflected in a higher MCE.
Our method exhibits a more even curve across bins and remains competitive in ECE while avoiding an explicit off-class KL penalty.

\paragraph{Calibration of predicted uncertainty}
We additionally test whether uncertainty is calibrated as a failure indicator.
Following the uncertainty-as-confidence view \cite{sirohi2022uncertaintyawarepanopticsegmentation,sirohi2022uncertaintyawarelidarpanopticsegmentation}, we define
$
\text{conf}_u(\vec x)=1-u_H(\vec x),
$
where $u_H\in[0,1]$ is the normalized predictive entropy of the expected class probabilities (Eq.~\eqref{eq:dir-total-entropy}).
We bin pixels by $\text{conf}_u$ and compute the micro-averaged uncertainty Expected Calibration Error (uECE).

Unlike $\text{conf}_p$, which depends only on the top class, $\text{conf}_u$ depends on the full predictive distribution.
Therefore, it is sensitive to residual probability mass on non-top classes.
A prediction can be correct and still receive a low $\text{conf}_u$ when the distribution is broad.
For this reason, the entropy-based reliability test is typically more demanding, and larger deviations from the diagonal are expected.
This behavior is visible in \autoref{fig:reliability-diagrams-conf-unc}(b).

The deterministic baseline shows a limited entropy dynamic range and remains overconfident under this view, with sparse occupancy in low-confidence bins.
MSE+KL becomes strongly underconfident in the entropy-based diagram, indicating that its predictive distributions are often too diffuse relative to the observed correctness frequencies.
NLL+KL and Digamma preserve roughly the same qualitative ordering seen in probability calibration, but their uncertainty calibration still degrades.
Our method yields the lowest uECE in this experiment.

Across the compared Dirichlet objectives, the KL regularizer is the primary factor shaping calibration behavior, but it can also worsen the calibration of distributional uncertainty by pushing predictions toward excessive dispersion or overly conservative confidence in parts of the range.
Our Invascal objective instead learns an input-dependent strength signal to control concentration directly, which yields competitive probability calibration and stronger uncertainty calibration under the entropy-based test.
\autoref{table:miouresults} summarizes results across all datasets and backbones.
Our method consistently ranks first or second in uECE, while ECE is more mixed relative to KL-regularized Dirichlet losses. Nevertheless, Invascal improves over the deterministic baseline (DET) in every test.

This shows that top-class ECE alone can be misleading: MSE+KL achieves the lowest ECE in our benchmark while exhibiting the worst uECE and lowest mIoU. Entropy-based calibration evaluates the full predictive distribution and therefore tests whether predictions are genuinely decisive rather than merely top-1 correct, making uECE a stronger reliability signal for risk-aware decision-making.


\begin{table}[!htbp]
\centering
\scriptsize
\setlength{\tabcolsep}{3pt}
\renewcommand{\arraystretch}{0.95}
\resizebox{\columnwidth}{!}{%
    \begin{tabular}{ l| cc |cc | cc}
    \hline
    & NLL & NLL+KL & Digamma & Digamma+KL & MSE & MSE+KL\\
    \hline
    ECE &
    4.8 &
    3.3 &
    4.7 &
    3.4 &
    3.8 &
    1.3 \\
    \hline
    \end{tabular}
}%
\renewcommand{\arraystretch}{1.0}
\setlength{\tabcolsep}{6pt}
\vspace{-0.2em} 
\caption{\small Expected Calibration Error (ECE) in percent for Dirichlet losses with and without KL regularizer on the SemanticKITTI validation set trained on model Reichert.}
\label{table:ECE_Dirichlet_w_wo_KL}
\end{table}

\subsection{Comparison with Temperature Scaling and MC Dropout}
\label{sec:exp_det_variants_vs_ours}

To contextualize our method against common calibration baselines, we compare Invascal with two deterministic post-/test-time alternatives built on the same SalsaNext DET model: (i) post-hoc Temperature Scaling (DET+TS) \cite{guo2017calibrationmodernneuralnetworks}, and (ii) MC Dropout \cite{SalsaNext} (DET+MC) as a sampling-based Bayesian approximation using 50 stochastic forward passes.
Results are reported in \autoref{table:abl}.

Invascal achieves the best aggregate calibration on this benchmark, with the lowest ECE and uECE, while maintaining high throughput.
DET+TS improves calibration over DET, especially for top-class confidence (ECE), but remains clearly behind Invascal in uncertainty calibration.
DET+MC is computationally expensive and substantially slower, while its calibration gains are mixed and do not match Invascal in uECE.
Overall, this experiment shows that Invascal provides a stronger calibration efficiency trade-off than both post-hoc temperature scaling and sampling-based MC Dropout.
\section{\large Conclusions and Future Work}
\label{sec:conclution}
Dirichlet evidential objectives can improve calibration over deterministic range-view LiDAR segmentation, but our experiments show that they may also be brittle in dense prediction and can incur large accuracy regressions depending on the loss and Kullback-Leibler regularization.
We proposed Invascal, a decoupled Dirichlet formulation that separates class preference from evidence strength and learns an input-dependent concentration control in a single forward pass.
Across three datasets and two backbones, Invascal preserves mIoU at the level of the deterministic baseline while consistently improving entropy-based uncertainty calibration (uECE) and remaining competitive on ECE, without relying on an explicit off-class KL penalty.
Future work could leverage the learned strength signal for risk-aware decision-making, active learning, and out-of-distribution perception in LiDAR range-view.

{\small
\bibliographystyle{IEEEtran} 
\bibliography{newbib}
}

\end{document}